# Exploring to establish an appropriate model for image aesthetic assessment via CNN-based RSRL: An empirical study

Ying Dai

Iwate prefectural university

dai@iwate-pu.ac.jp

**Abstract**

To establish an appropriate model for photo aesthetic assessment, in this paper, a D-measure which reflect the disentanglement degree of the final layer FC nodes of CNN is introduced. By combining F-measure with D-measure to obtain a FD measure, an algorithm of determining the optimal model from the multiple photo score prediction models generated by CNN-based repetitively self-revised learning (RSRL) is proposed. Furthermore, the first fixation perspective (FFP) and the assessment interest region (AIR) of the models are defined and calculated. The experimental results show that the FD measure is effective for establishing the appropriate model from the multiple score prediction models with different CNN structures. Moreover, the FD-determined optimal models with the comparatively high FD always have the FFP an AIR which are close to the human's aesthetic perception when enjoying photos.

## 1. Introduction

With the great growth of digital pictures, many researches have been interested in exploring the methods of image aesthetic auto-assessment. However, the research on this field is challenging due to the subjectivity and ambiguity of aesthetic criteria, and the imbalance of the quality distribution. In [1], the authors give an experimental survey about this field's research. In this paper, besides the discussion of main contributions of reviewed approaches, the authors systematically evaluate deep leaning settings that are useful for developing a robust deep model for aesthetic scoring. Moreover, they discuss the possibility of manipulating the aesthetics of images through computational approaches. Recently, facing the issues of the subjectivity and ambiguity of aesthetic criteria, besides predicting the mean opinion score provided by data sets, the approach of predicting the distribution of human opinion scores using a convolutional neural network (CNN) is proposed [2]. Further, the authors use the proposed assessment technique to effectively tune parameters of image denoising and tone enhancement operators to produce perceptually superior results. However, due to the restriction of the CNN, all images are rescaled to square images to feed into the network regardless of their aspect ratios. Following the work in [2], Lijie Wang et al. propose a method of aspect-ratio-preserving multi-patch image aesthetics score prediction is proposed [3], in order to reflect the original aspect ratio information to prediction. In [4], besides keeping the original aspect ratio, the authors propose a spatial attentive image aesthetic assessment model to evaluate image layout, and find spatial importance in aesthetics. In order to improve the learning efficiency during the training process, a multi-patch aggregation method for image aesthetic assessment with preserving the original aspect ratio is proposed [5]. In this method, the goal is achieved by resorting to an attention-based mechanism that adaptively adjusts the weight of each patch of the image. Moreover, in [6], authors propose a gated peripheral-foveal convolutional neural network. It is a double-subnet neural network. The former aims to encode the holistic information and provide the attended regions. The latter aims to extract fine-grained features on these key regions. Then, a gated information fusion network is employed for the image aesthetic prediction. In [7], the authors propose a novel multimodal recurrent attention CNN, which incorporates the visual information with the text information. This method employs the recurrent attention network to focus on some key regions to extract visual features. However, it has been validated that feeding the weighted key regions to CNN to train the image aesthetic assessment model degrades the performance of prediction according to our preliminary experiments, because the aesthetic assessment is effected by holistic information in the image. Weakening some regions results in the information degradation for aesthetic assessment.

Furthermore, these methods focus on training the model to predict the distribution of human opinion scores so as to obtain the mean score of the image. So, the personal opinion score that is important in the aesthetic assessment is not reflected. In [8], authors propose a unified algorithm to solve the three problems of image aesthetic assessment, score regression, binary classification, and personalized aesthetics based on pairwise comparison. The model for personalized regression is trained on FLICKERAES dataset [9]. However, the ground truth score was set to the mean of five workers' scores. Accordingly, whether the predicted score embodies the inherently personal aesthetics is not clear.

Moreover, the above papers ignore the fact that the distribution of samples against scores in dataset is highly non-uniform [10] [11]. Most images in the dataset assessed by a professional photographer have the score of 4, (about 45%), and about 85% of the images in the dataset concentrate on the scores of 3 to 5 (about 87%). The score classification model could be overwhelmed by those samples in the majority classes if the parameters are learned by treating all samples equally. Accordingly, in order to solve the problem of the imbalanced classification in image aesthetic assessment, the papers of [10] [11] propose a mechanism of repetitively self-revised learning (RSRL) to train the CNN-based image score prediction model on the imbalanced score data set. As RSRL, the neural networks are trained repetitively by dropping out the low likelihood photo samples at the middle levels of aesthetics from the training data set based on the previously trained network, and the optimal model that have the highest value of F-measure is selected from them. The experimental result presents that this model outperforms the model trained without RSRL. However, because the F-measure is dependent on the test dataset, whether it reflect the internal property of the model is not clear.

On the other hand, some researchers aim at extracting and analyzing the aesthetic features to find the relation with the aesthetic assessment. In [12], the paper presents in-depth analysis of the deep models and the learned features for image aesthetic assessment in various viewpoints. In particular, the analysis is based on transfer learning among image classification and aesthetics classifications. The authors find that the learned features for aesthetic classification are largely different for those for image classification, i.e., the former accounts for color and overall harmony, while the latter focuses on texture and local information. However, whether this finding is universal needs to be validated further. In [13], besides extracting deep CNN features, five algorithms for handcrafted extracting aesthetic feature maps are proposed, which are used to extract feature maps of the brightness, color-harmony, rule of thirds, shallow depth of filed, and motion blur of the image. Then, a novel feature fusion layer is designed to fuse aesthetic features and CNN features to improve the aesthetic assessment. However, the experimental result shows that the fusion only improves the accuracy of 1.5% over no-fusion. Accordingly, whether it is necessary to incorporate the inefficiently handcrafted aesthetic features with deep CNN features is needed to investigate.

In this paper, we aims at exploring to establish an appropriate model for image aesthetic assessment via CNN-based RSRL on the imbalanced score dataset, and verify the adaptability to the other out-of-distribution (o.o.d) dataset. Further, we try to acquire first fixation perspective and assessment interest region from the learned feature maps, so as to make the established model interpretable in what aesthetic features are learned and whether these reflect the aesthetic attributes when human enjoying photos.

In details, the main contributions of this paper are summarized as follows.

- For RSRL, besides the external F-measure, an internal measure called disentanglement-measure, which measuring the degree of disentanglement of final FC layer nodes, is defined. The experimental results shows that the aggregation of F-measure and disentanglement-measure is helpful for determining the optimal model from the models retrained by RSRL.
- Three kinds of CNN-based models, which are fine-tuned AlexNet, FC layers-added AlexNet, and only 1x1 convolutions-used CNN, are trained on xiheAA [10]. Experimental results shows that only 1x1 convolutions-used CNN rivals AlexNet on CUHK-PQ [1] of o.o.d, although FC layers-added AlexNet outperforms other two models 10% on F-measure. However, the size of 1x1 convolutions-used CNN is only about 1/620 of FC layers-added AlexNet and AlexNet.
- The first fixation perspective and the assessment interest region are defined and acquired from the learned feature maps. It is found that those acquired from the FC layers-added AlexNet and 1x1 convolutions CNN seem to be more close to the human's aesthetic perception.

The reminder of this paper is organized as follows. Section 2 introduces a disentanglement measure and Section 3 describes the mechanism of establishing an appropriate model for image aesthetic assessment via RSRL. Section 4 explains the method of extracting the first fixation perspective and the assessment interest region of the model. Section 5 gives the experimental results and analyze the effectiveness of the proposed methods.

## 2. Disentanglement measure

The approach of RSRL is proposed in [10]. In this paper, in order to solve the data imbalance in training the model for photo aesthetic assessment, the author focuses on conducting repetitive self-revised learning (RSRL) to retrain the CNN-based photo score prediction model repetitively by transfer learning, so as to improve the performance of imbalanced classification caused by the highly non-uniform distribution of training samples against scores. As the repetitive self-revised learning, the network is trained repetitively by dropping out the low likelihood photo samples with mid-level scores from the training data set based on the previously trained model. Then, as the photo score predictor, the optimal model is determined from themultiple models according to the F-measure. However, F-measure is an external measure of the model. Whether it reflects the intrinsic property of the model is unclear. Accordingly, this paper conducts a new internal measure called disentanglement-measure which measures the degree of disentanglement of final FC layer nodes of CNN. The idea behind the disentanglement-measure is that the nodes of the final FC layer should be disentangled for the accurate classification. However, it is obvious that the adjacent score classes are relevant. So, it is impossible for score classification to disentangle the final FC layer nodes of CNN completely. Then, the disentanglement-measure of these nodes against their weights is defined. In the following, we describe how to calculate the disentanglement-measure in details.

Fig. 1 shows the situation of the last two FC layers of CNN.

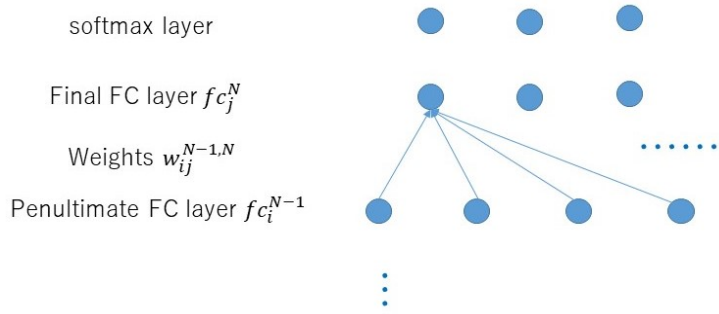

Fig.1 Last two FC layers of CNN

Let the node of the final FC layer be $fc_j^N$, the node of the penultimate FC layer be $fc_i^{N-1}$, and the weight between $fc_i^{N-1}$ and $fc_j^N$ be $w_{ij}^{N-1,N}$. Further, the number of $fc_j^N$ is $J$, and the number of $fc_i^{N-1}$ is $I$. The disentanglement-measure is calculated according to the following steps.

- Normalizing $w_{ij}^{N-1,N}$ as $w1_{ij}$;
- Calculating the correlation matrix of the final FC layer nodes based on $w1_{ij}$;
  $R = \frac{1}{I-1} w1'_{ij} * w1_{ij}$ , i ∈ [1, I], and j ∈ [1, J]     (1)
  R is a matrix of $J \times J$.
- Calculating and sorting the eigenvalues in a descent order, denoted $eign_m$ and obtaining the corresponding eign-vectors, denoted $eign\_v_m$;
- Calculating the factor loading of factor $m$ (latent variable) against $fc_j^N$, j∈[1,J];
  $fl_{m,j} = \sqrt{eign_m} * eign\_v_{m,j}$, m ∈ [1, M]     (2)
  M indicates the number of factors.
- Calculating the 2-norm of factor loadings against the two nodes *j1* and *j2* of the final FC layer;
  $dis_{j1,j2} = 2 - norm(fl_{m,j1} - fl_{m,j2})$     (3)
  The larger value indicate that the two nodes *j1* and *j2* are more leaved.
- Calculating the minimum of $dis_{j,jj}$ regarding one node *j* to all of the other nodes

$jj$ of the final FC layer;

$$dis\_min_j = \min(dis_{j,jj}) \quad (4)$$

- Calculating the mean of $dis\_min_j$ regarding all of the nodes of the final FC layer;

We can see that the mean of $dis\_min_j$ ($j \in J$) reflect the degree of dispersion of the nodes according to the above calculation. The larger value indicates that the nodes are more scattered. So, we define the disentanglement measure (D-measure) of the nodes of the final FC layer as the following expression.

$$D - measure = mean(dis\_min_j) \quad (5)$$

In the next section, the role of D-measure in establishing an appropriate model for image aesthetic assessment via CNN-based RSRL will be discussed.

## 3. Establishing an appropriate model for image aesthetic assessment via RSRL

An improved mechanism of CNN-based RSRL is shown in Fig.2.

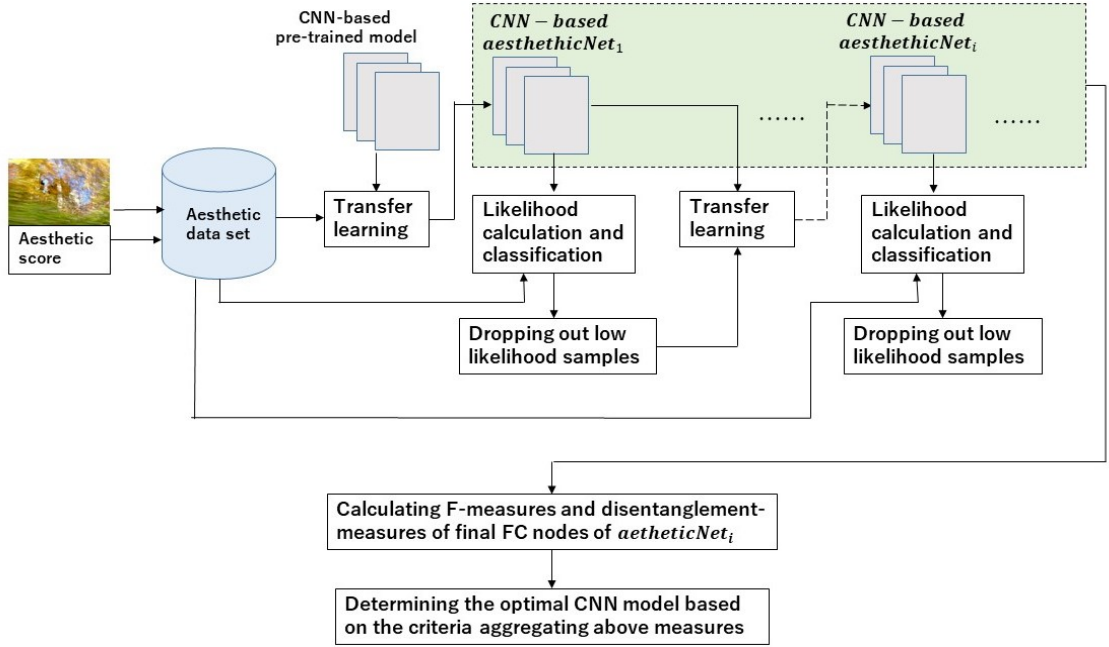

Fig.2 improved CNN-based RSRL

The mechanism of CNN-based RSRL is explained in [10] [11]. The approach of CNN-based RSRL is to drop out the low likelihood samples of majority classes of scores repetitively, so as to ameliorate the invasion of these samples to the minority classes, and prevent the loss of the samples with discriminative features in the majority classes. In this process, the previous model are re-trained by transfer learning again and again. Accordingly, many re-trained models are generated with RSRL. Then, the optimal model is determined among these models based on the F-measures. However, the F-measure is dependent on the test dataset reflecting the external property of the models, so, whether it embody the internal property of the model is not clear. In this paper, we introduce D-measure with F-measure to determine the optimal model because D-measure reflect the internal property of the model. In details, let the F-measure of class $j$ be $F_j$, and the total F-measure of the classes be F_all. F_all is calculated by the equations (6).

$$\text{F\_all} = \sum_{j=1}^{J} F_j \qquad (6)$$

Then, the measure that aggregates F_all and D-measure is defined by the equation (7). Here, the values of F_all and D-measure are normalized.

$$\text{FD} = w_1 F\_all + w_2 D - measure \qquad (7)$$

Where, $w_1$ and $w_2$ are set as 0.5, respectively.

It is assumed that the optimal model should be one that has the maximal value of F_all among the re-trained models, and the value of its FD is larger than a threshold $T$. Then, the algorithm selecting the optimal model is expressed in the expression (8).

$$L_{optimal} = \begin{cases} \underset{l \in L}{argmax} \; F\_all_l \\ FD_{L_{optimal}} > T \end{cases} \qquad (8)$$

Where, the $L_{optimal}$ indicates the index of the optimal model, and $l$ and $L$ indicate the index and the number of the re-trained models, respectively. In this paper, $T$ is set to 0.95, which will be explained in subsection 5.1.

On the other hand, if the F_all or the FD is only used to determine the optimal model, the indices of the selected models are expressed by the expression (9) and (10), respectively.

$$L_{optimal}^{F\_all} = \underset{l \in L}{argmax} \; F\_all_l \qquad (9)$$

$$L_{optimal}^{D} = \underset{l \in L}{argmax} \; D_l \qquad (10)$$

Then, scores of images can be predicted by the aggregation of models $L_{optimal}^{F\_all}$ and $L_{optimal}^{D}$. The formulation for predicting the score is expressed by the equation (11)

$$\text{score} = \underset{j \in J}{argmax}(w_1 fc_j^{F\_all} + w_2 fc_j^{D}) \qquad (11)$$

Where, $fc_j^{F\_all}$ and $fc_j^{D}$ indicate the sigmoid values of the final FC layers regarding the models $L_{optimal}^{F\_all}$ and $L_{optimal}^{D}$, respectively. $w_1$ and $w_2$ are set as 0.5, respectively.

The effectiveness of conducting D-measure and model $L_{optimal}^{D}$ in the prediction is validated by experiments. The results and analysis will be shown in section 5.

## 4. Extracting first fixation perspective and assessment interest region

Although there are composition attributes for taking good photos, such as rule of third and depth of field, people are more likely to concern the first fixation perspective (FFP) and the relation with other elements when enjoying photos, which are considered to be the assessment interest region (AIR). Accordingly, for a CNN-based photo aesthetic assessment model, it is supposed that the most activated feature map should be related to the FFP of the image, and the sum of feature maps should be related to the AIR. So, the model's FFP and AIR could be acquired by the following calculation.

- obtaining the most activated feature map and the sum of feature maps of the final convolutional layer of CNN regarding an image I(x,y), and normalizing and resizing these to the size equal to the I, where, x and y indicate the coordinates of the pixel, respectively;

Let a feature map of the final convolutional layer be $FM^p(x,y)$, its index is $p$, the number of feature maps is P, and the index of the most activated feature map be $P_{max}$.

$$P_{max} = \underset{p \in P}{\mathrm{argmax}}\, FM^p(x,y) \qquad (12)$$

Moreover, let the most activated feature map be $FM_{max}(x,y)$, and the sum of feature maps be $FM_{sum}(x,y)$.

$$FM_{sum}(x,y) = \sum_{p=1}^{P} FM^p(x,y) \qquad (13)$$

- Extracting first fixation perspective and assessment interest region;

Let the first fixation perspective be FFP(x,y), and assessment interest region be AIR(x,y).

$$\mathrm{FFP}(x,y) = \mathrm{I}(x,y) * FM_{max}(x,y) \qquad (14)$$

$$\mathrm{AIR}(x,y) = \mathrm{I}(x,y) * FM_{sum}(x,y) \qquad (15)$$

Fig.3 shows the examples of FFP(x, y) and AIR(x, y) of an image, which are calculated based on the learned feature maps of the model with three 1x1 convolution layers.

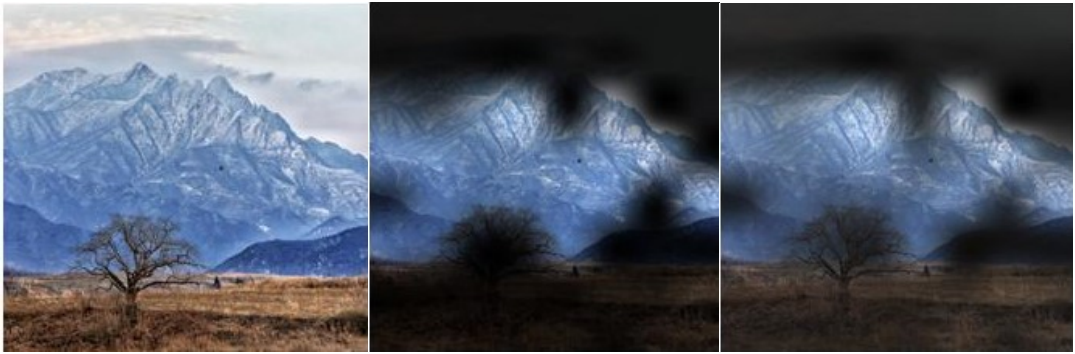

(a) Original image  (b) FFP(x, y)  (c) AIR(x, y)

Fig 3 examples of FFP(x, y) and AIR(x, y)

It seems that the high light regions of (b) and (c) are close to the human's perception regarding the first fixation perspective and the aesthetic assessment interest region when enjoying an image.

In the section 5, we will show whether the optimal model determined by (8) could extract the FFP(x, y) and AIR(x, y) which are close to the human's aesthetic perceptions.

## 5. Experimental results and analysis
### 5.1 Establishing the appropriate model

In this paper, we construct three kinds of CNN models with different structures to validate how the appropriate model can be determined by expression (8). These three kinds of CNN structures are fine-tuned AlexNet, changed AlexNet, and new designed only 1x1 convolutions CNN, which are presented below.

Type (a) Fine-tuned AlexNet

*1--end-3 layers:   transferring 1--end-3 of AlexNet*

*end-2 layer 'fc': 8 fully Connected layer, each corresponding to a score class of 2~9*

*end-1 layer 'softmax': Softmax*

*end layer 'classoutput': Classification Output*

Type (b) Changed AlexNet

*1--end-9 layers: transferring 1--end-5 of AlexNet*

*end-8 layer 'batchnorm_1': Batch normalization with 4096 channels*

*end-7 layer 'relu_1': ReLU*

*end-6 layer 'dropout': 50% dropout*

*end-5 layer 'fc_1': 32 fully connected layer*

*end-4 layer 'batchnorm_2': Batch normalization with 32 channels*

*end-3 layer   'relu_2': ReLU*

*end-2 layer 'fc_2': 8 fully Connected layer, each corresponding to a score class of 2~9*

*end-1 layer 'softmax': Softmax*

*end layer 'classoutput': Classification Output*

Type (c) Only 1x1 convolutions' CNN

*1 layer 'imageinput': 227×227×3 images with 'zerocenter' normalization*

*2 layer 'conv_1': 94 1×1×3 convolutions with stride [8   8] and padding [0   0   0   0]*

*3 layer 'batchnorm_1': Batch normalization with 94 channels*

*4 layer 'relu_1': ReLU*

*5 layer 'conv_2': 36 1×1×94 convolutions with stride [4   4] and padding [0   0   0   0]*

*6 layer 'batchnorm_2': Batch normalization with 36 channels*

*7 layer 'relu_2': ReLU*

*8 layer 'conv_3': 36 1×1×36 convolutions with stride [1  1] and padding [0  0  0  0]*

*9 layer 'batchnorm_3': Batch normalization with 36 channels*

*10 layer 'relu_3': ReLU*

*11 layer 'fc_1': 36 fully connected layer*

*12 layer 'fc_2': 8 fully Connected layer, each corresponding to a score class of 2~9*

*13 layer 'softmax': Softmax*

*14 layer 'classoutput': Classification Output*

By using transfer learning, each of these CNN are re-trained 29 times iteratively via RSRL on xiheAA dataset [10], which was mentioned in the section introduction. 4/5 samples are randomly selected as the training dataset, and the remained 1/5 samples are served as the validation dataset. Fig. 4, Fig. 5 and Fig.6 show the values of D-measure, F_all, and FD of 29 re-trained models regarding three kinds of CNNs on the validation dataset, respectively. The values of D-measure and F_all are normalized.

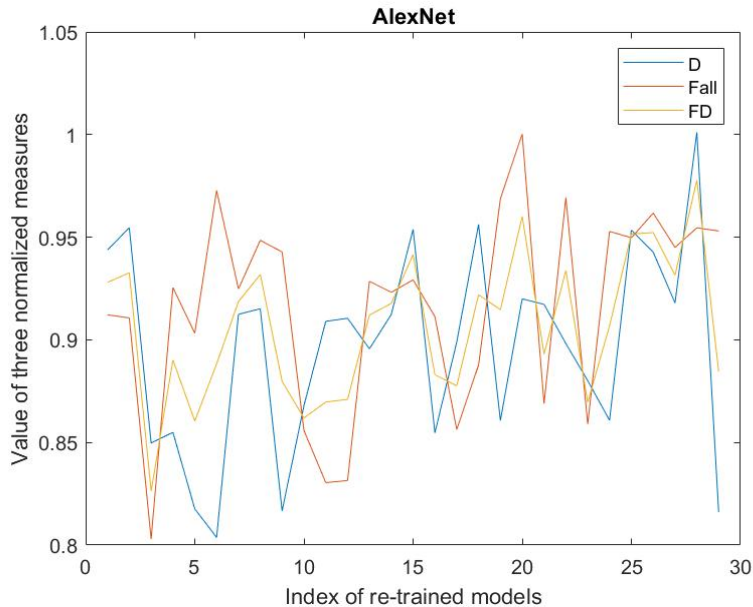

Fig. 4 Results regarding fine-tuned AlexNet

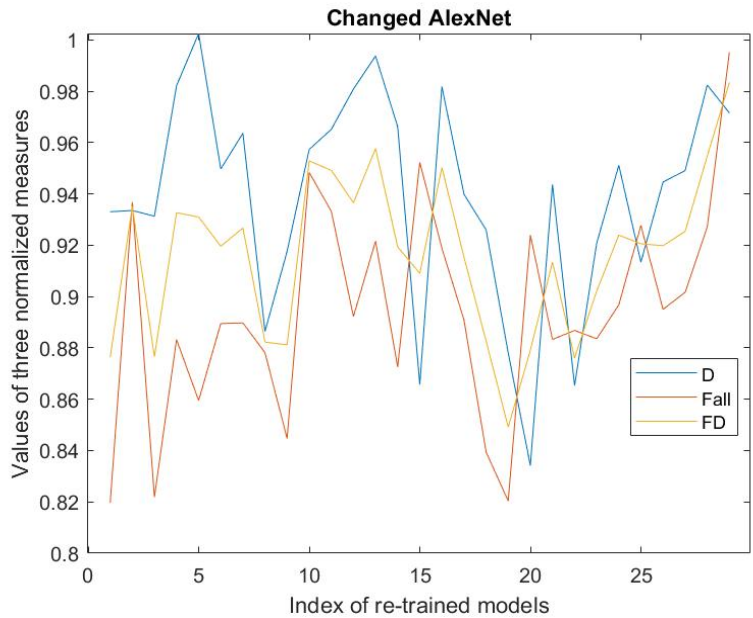

Fig. 5 Results regarding changed AlexNet

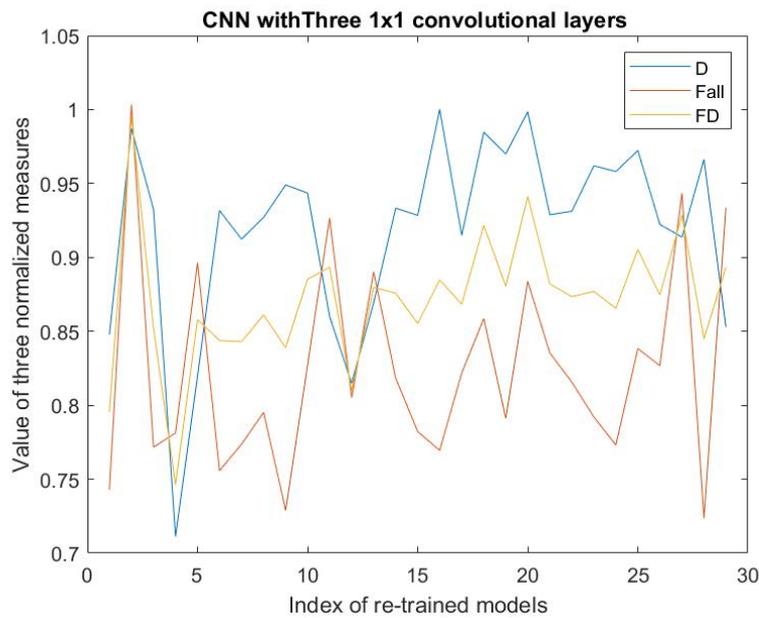

Fig.6 Results regarding only 1x1 convolutions' CNN

From the results shown in these figures, we can see that for the models of changed AlexNet (type (b)) and CNN with three 1x1 convolutional layers (type (c)), the optimal models with maximal FD are the models with maximal F_all. Simultaneously, these models have the comparatively high values of D-measure. The FD measures can reach the values more than 0.98.

For the model of fine-tuned AlexNet (type (a)), although the re-trained model with maximal FD is the model with maximal D-measure, the model of index 20 having second largest FD is the one with maximal F_all. Moreover, the FD of that is 0.96, larger than the threshold 0.95. So, this model should be selected as the optimal model.

On the other hand, to the various re-trained models, we notice that the models with high D-measure values are not necessarily ones with high F_all values; vice versa.

Based on the above results and analysis, we can say that the F_all measuring the external performance of the model could reflect the internal property of the classification. It is possible to use the maximal F_all to select the optimal model from the re-trained models. The D-measure measuring the internal classification property of the model is a necessary condition for the good classification, but it is not a sufficient condition. Accordingly, it is better to use the FD measure that aggregate the F_all and the D-measure to assist on obataining the optimal model. If the value of FD reaches a threshold, for example, 0.95, RSRL can be stopped, and the current re-trained model is used as the optimal model. The benefits of doing so can improve the efficiency of RSRL. For example, in the case of Fig.6, RSRL can be stopped at second iteration, while the FD is larger than 0.95.

Accordingly, trained on xiheAA dataset, the model of type (a) with index 20 (Alex20), the model of type (b) with index 29 (CAlex29), and the model of type (c) with index 2 (1x1CNN2) are selected as the optimal models, respectively. The values of FD is 0.96, 0.98 and 0.995, respectively.

Next, the CUHK-PQ dataset [1] is used to the out-of-distribution validation for the above selected models. The CUHK-PQ dataset contains 10524 high quality images and 19166 low quality images. So, the images predicted having the score less than 5 are assigned to the low class, the others are assigned to the high class. Fig. 7 shows the experimental results.

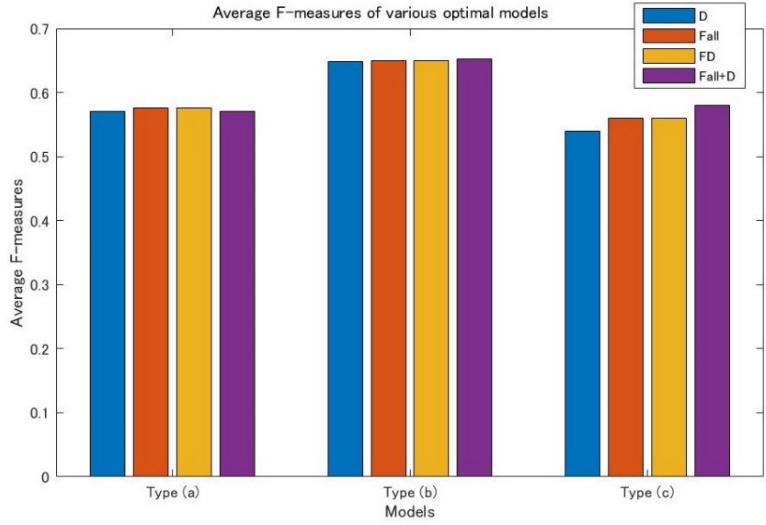

Fig. 7 Average F-measures of various optimal models

The average F-measures of various optimal models regarding type (a), (b) and (c) are presented in the fig.7. The yellow bars indicate the values of Alex20, CAlex29, and 1x1CNN2, respectively. The red bars and blue bars indicate the values of the optimal models determined by F_all and D-measure. It is obvious that the heights of red bars and yellow bars are same, but the heights of blue bars are little low although those are very close to the yellow bars. The purple bars indicate the values of the model generated by (11). It is observed that the purple height of type (a) is slightly lower than the yellow bar; that of (b) is slightly higher than the yellow bars; but, that of type (c) are obviously higher than the yellow bar. Because the FD values of Alex20, CAlex29, and 1x1CNN2 indicated by yellow bars are 0.96, 0.98 and 0.995, respectively, it seems that the higher the FD, the better the result is in the aggregation $L_{optimal}^{F\_all}$ and $L_{optimal}^{D}$.

As a whole, the above analyses indicate further that the D-measure is a necessary condition for the good classification, but it is not a sufficient condition. It is better to use the FD measure that combines the F_all and the D-measure in selecting the optimal models. Especially, if the FD-determined optimal model has the comparatively high FD, introducing $L_{optimal}^{D}$ to aggregate $L_{optimal}^{F\_all}$ can improve the performance of the score prediction obviously.

On the other hand, it is observed that the models of type (b) have about 10% higher average F-measures than the other two types of models. That is, inserting new layers into the FC section of CNN can improve the performance of predicting scores. However, it is more interested that the average F-measure of the optimal model of type (c) determined by FD is almost as same as that of type (a). Moreover, the average F-measure of type (c) obtained by the aggregation of models $L_{optimal}^{F\_all}$ and $L_{optimal}^{D}$ is 1.8% higher than that of type (a). Accordingly, we could deduce that the new designed CNN with only three layers 1x1 convolutions via RSRL on a small dataset could outperforms the fine-tuned pre-trained CNN, such as AlexNet. Moreover, the size of type (c) is only about 1/620 of type (a) and type (b). So, in the view of the ratio of cost-effectiveness, the only 1x1 convolutions CNN may be better as a score prediction model, although the type (b) which is the changed AlexNet is better in the view of F-measure.

### 5.2 Extracting first fixation perspective and assessment interest region

Fig.9 and Fig 10 show the examples of first fixation perspectives (FFP(x, y)) and assessment interest regions (AIR(x, y)) of an image shown in Fig.8, which are acquired based on (14) and (15) using the FD-determined and the D-determined optimal models of type (a), (b) and (c), respectively.

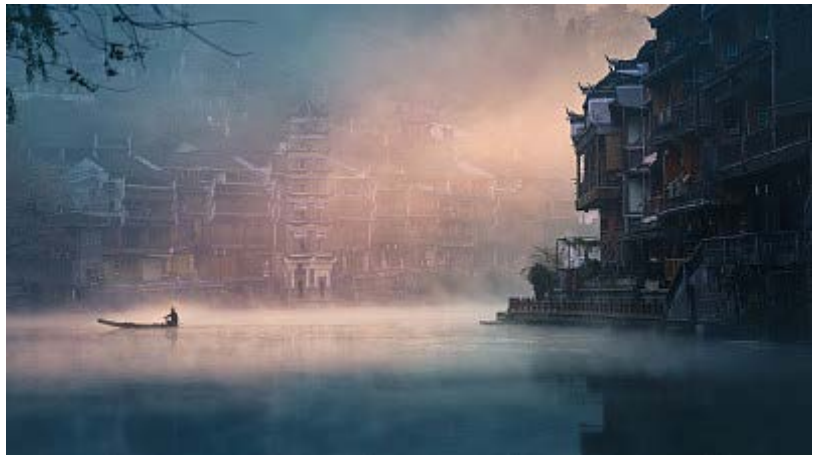

Fig. 8 Original image

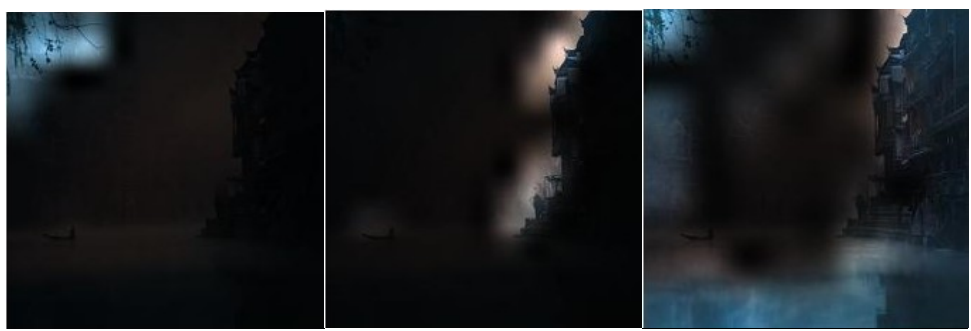

Type (a)         Type (b)         Type (c)

FFP(x, y) regarding the FD-determined optimal model

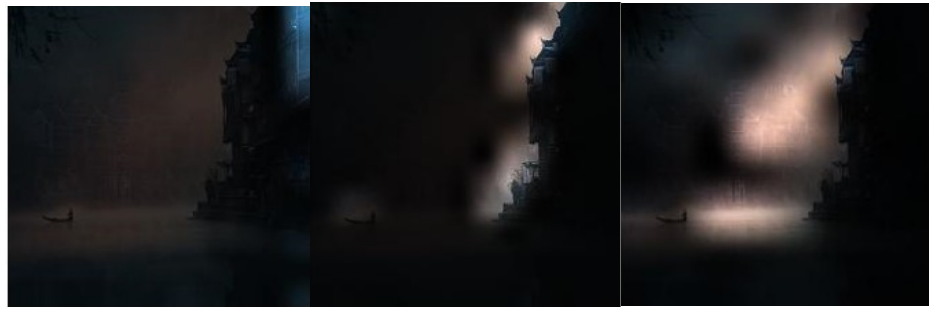

| Type (a) | Type (b) | Type (c) |

FFP(x, y) regarding the D-determined optimal model

Fig. 9 First fixation perspective

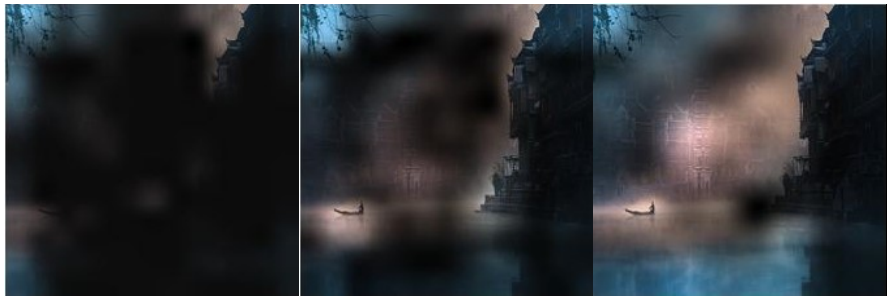

| Type (a) | Type (b) | Type (c) |

AIR(x, y) regarding the FD-determined optimal model

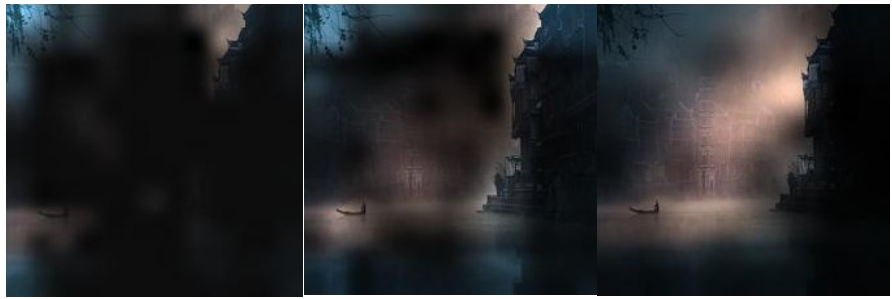

| Type (a) | Type (b) | Type (c) |

AIR(x, y) regarding the D-determined optimal model

Fig. 10 Assessment interest region

The scores predicted by the FD-determined optimal models of types (a), (b) and (c) are 6, 6, and 3, respectively; those by D-measure are 6, 6, and 6 respectively. The score predicted by (11) is 7.

From the results, we can see that for all of the types, two kinds of AIR(x, y) are almost same. Especially, in the case of types (b) and (c), the high light region that is considered as the assessment interest region is the light orange area, while it is very close to the human's perception when enjoying the photo. However, in the case of type (a), the extracted assessment interest region is the surrounding area of the photo that seems to diverge the human's perception, although the predicted score seems to be right. These observations maybe indicate that fine-tuning pre-trained CNN on the small score dataset can't make the re-trained model learn the deep aesthetic features; but, it may be possible to make the CNN models learn the deep aesthetic features by changing pre-trained CNN structure or training new multi-layer only 1x1 convolutions CNN.

On the other hand, the two kinds of FFP(x, y) are not same for the models of type (a) and type (c), although they are same for the model of (b). We think that the FFP of the image should be the light orange area for the human's aesthetic perception. The two optimal models of type (b) indeed learn the elements of FFP; however, in the case of (c), FD-determined optimal model extract the background of the light orange area as FFP, although the D-determined optimal model learn the light orange area. Maybe this is a reason for that model to assign a score 3 to the photo. In the case of (a), we think that the areas extracted by the two models as FFP are far from the human's visual perception of that. So, these observations confirm the above statement further that it is difficult to fine-tune pre-trained CNN to make the re-trained model learn the deep aesthetic features; but, it may be possible to make the CNN models learn the deep aesthetic features by changing pre-trained CNN structure or training new multi-layer only 1x1 convolutions CNN.

As a whole, we can say that extracting FFP and AIR of the models can help in understanding the internal properties of the models. The FD-determined optimal models with the comparatively high FD always have the FFP an AIR which are close to the human's aesthetic perception.

## 6. Conclusion

To establish an appropriate model for photo aesthetic assessment, in this paper, a D-measure which reflect the disentanglement degree of the final layer FC nodes of CNN was introduced. By combining F-measure with D-measure to obtain a FD measure, an algorithm of determining the optimal model from the multiple photo score prediction models generated by CNN-based RSRL was proposed. Furthermore, the first fixation perspective (FFP) and the assessment interest region (AIR) of the models were defined and calculated. By the experimental results, we can say that introducing D-measure is really helpful in selecting the optimal model from the re-trained models. The proposed algorithm is effective for establishing the appropriate model from the multiple photo score prediction models with different CNN structures. If the FD-determined optimal model has the comparatively high FD, introducing $L_{optimal}^{D}$ to aggregate $L_{optimal}^{F\_all}$ can improve the performance of the score prediction obviously. Furthermore, obtaining FFP and AIR of the models can help in understanding the internal properties of these models. The FD-determined optimal models with the comparatively high FD always have the FFP an AIR which are close to the human's aesthetic perception when enjoying photos.

## Reference

[1] Yubin, Chen Change Loy, and Xiaoou Tang, "Image aesthetic assessment: an experimental survey", IEEE Signal Processing Magazine, Volume 34, Issue 4, Pages 80-106, July 2017.

[2] Hossein Talebi and Peyman Milanfiar, "NIMA: Neural image assessment", IEEE Trans. on image processing, Vol. 27, No. 8, Pages 3998-4011, Aug. 2018.

[3] Lijie Wang, et al. "Aspect-ratio-preserving multi-patch image aesthetics score prediction", Proceedings of the IEEE/CVF Conference on Computer Vision and Pattern Recognition (CVPR) Workshops, 2019.

[4] Ying Xu, et al. "Spatial attentive image aesthetic assessment", Proceedings of 2020 IEEE international conference on Multimedia and expo (ICME), DOI: 10.1109/ICME46284.2020.9102804, 2020.

[5] Kekai Sheng, et al. "Attention-based multi-patch aggregation for image aesthetic assessment", Proceedings of ACM'MM, pages 879-886, 2018.

[6] Xiaodan Zhang, et al. "A gated peripheral-foveal convolution neural network for unified image aesthetic prediction", IEEE Trans. on multimedia, Vol. 21, Issues 11, Pages: 2815-2826, Nov. 2019.

[7] Xiaodan Zhang, et al. "Beyond vision: A multimodal recurrent attention convolutional neural network for unified image aesthetic prediction tasks", IEEE Trans. on multimedia, Vol. 23, pp. 611-623, Apr. 2020.

[8] Jun-Tae Lee and Chang-Su Kim, "Image Aesthetic Assessment Based on Pairwise Comparison – A Unified Approach to Score Regression, Binary Classification, and Personalization", Proceedings of the IEEE/CVF International Conference on Computer Vision (ICCV), 2019, pp. 1191-1200.

[9] Jian Ren, et al. "Personalized image aesthetics", Proceedings of 2017 IEEE International Conference on Computer Vision (ICCV), DOI: 10.1109/ICCV.2017.76, 2017.

[10] Ying Dai, "Sample-specific repetitive learning for photo aesthetic auto-assessment and highlight elements analysis", Multimedia Tools and Applications, Vol. 80, pp. 1387–1402, DOI 10.1007/s11042-020-09426-z, 2021.

[11] Ying Dai, "CNN-based repetitive self-revised learning for photos' aesthetics imbalanced classification", Proc. of IEEE 25th International Conference on Pattern Recognition (ICPR), pp. 331-338, 2021.

[12] Hyeongnam Jiang and Jong-Seok Lee, "Analysis of deep features for image aesthetic assessment", IEEE Access, Vol. 9, pp. 29850-29861, 2021.

[13] Xuewei Li, et al. "A novel feature fusion method for computing image aesthetic quality", IEEE access, Vol. 8, pp. 63043-63054, 2020